\documentclass[11pt,a4paper]{article}

\usepackage{geometry}
\usepackage{times}
\usepackage{latexsym}
\usepackage{url}
\usepackage{booktabs}       
\usepackage{amsfonts}       
\usepackage{nicefrac}       
\usepackage{microtype}      

\usepackage{algorithmic}
\usepackage{algorithm}
\usepackage{wrapfig}
\usepackage{lipsum}
\usepackage{xspace}
\usepackage{amsmath}
\usepackage{bm}
\usepackage{graphicx} 
\usepackage{subcaption}
\usepackage[skip=5pt]{caption}
\usepackage{paralist}
\usepackage{float}
\usepackage{tabularx}
\usepackage{multirow}
\usepackage{diagbox}
\usepackage{tikz}
\usepackage{booktabs}
\usepackage{array, makecell} 
\usepackage{natbib}
\usepackage{amsmath,amssymb}
\usepackage[inline]{enumitem}
\usepackage{acronym}
\usepackage[colorlinks,allcolors=violet]{hyperref}

\newcounter{todocnt}

\newcommand{\name}{IGLU\xspace}

\acrodef{RL}{Reinforcement Learning}
\acrodef{RNN}{Recurrent Neural Net}

\title{NeurIPS 2021 Competition IGLU:\\
Interactive Grounded Language Understanding \\ in a Collaborative Environment}
\author{Julia Kiseleva\textsuperscript{1}       \and
        Ziming Li\textsuperscript{3}            \and  
        Mohammad Aliannejadi\textsuperscript{2} \and 
        Shrestha Mohanty\textsuperscript{1}     \and 
        Maartje ter Hoeve\textsuperscript{2}    \and 
        Mikhail Burtsev\textsuperscript{4,5}    \and  
        Alexey Skrynnik\textsuperscript{4}      \and 
        Artem Zholus\textsuperscript{4}         \and
        Aleksandr Panov\textsuperscript{4}      \and 
        Kavya Srinet\textsuperscript{6}         \and 
        Arthur Szlam\textsuperscript{6}         \and
        Yuxuan Sun\textsuperscript{6}           \and 
        Katja Hofmann\textsuperscript{1}        \and
        Michel Galley\textsuperscript{1}        \and
        Ahmed Awadallah\textsuperscript{1} \\
        \\
\textsuperscript{1}Microsoft Research,  
\textsuperscript{2}University of Amsterdam, \\
\textsuperscript{3}Alexa AI, 
\textsuperscript{4}MIPT,
\textsuperscript{5}AIRI,
\textsuperscript{6}Facebook AI
}

\begin{document}
\maketitle

\begin{abstract}

Human intelligence has the remarkable ability to quickly adapt to new tasks and environments. Starting from a very young age, humans acquire new skills and learn how to solve new tasks either by imitating the behavior of others or by following provided natural language instructions. To facilitate research in this direction, we propose \emph{IGLU: Interactive Grounded Language Understanding in a Collaborative Environment}.

The primary goal of the competition is to approach the problem of how to build interactive agents that learn to solve a task while provided with grounded natural language instructions in a collaborative environment. Understanding the complexity of the challenge, we split it into sub-tasks to make it feasible for participants.  

This research challenge is naturally related, but not limited, to two fields of study that are highly relevant to the NeurIPS community: Natural Language Understanding and Generation (NLU/G) and Reinforcement Learning (RL). 
Therefore, the suggested challenge can bring two communities together to approach one of the important challenges in AI. Another important aspect of the challenge is the dedication to perform a human-in-the-loop evaluation as a final evaluation for the agents developed by contestants.
\end{abstract}

\subsection*{Keywords}
Natural Language Understanding (NLU), Reinforcement Learning (RL), Grounded Learning, Interactive Learning, Games
\subsection*{Competition type} Regular

\section{Competition description}

\subsection{Background and impact}
Humans have the remarkable ability to quickly adapt to new tasks and environments. Starting from a very young age, humans acquire new skills and learn how to solve new tasks either by imitating behavior of others or by following natural language instructions that are provided to them~\citep{an_imitation_1988, council_how_1999}. 
Natural language communication provides a natural way for humans to acquire new knowledge, enabling us to learn quickly through language instructions and other forms of interaction such as visual cues. This form of learning can even accelerate the acquisition of new skills by avoiding trial-and-error and statistical generalization when learning only from observations~\citep{thomaz2019interaction}. Studies in developmental psychology have shown evidence of human communication being an effective method for transmission of generic knowledge between individuals as young as infants~\citep{csibra2009natural}. These observations have inspired attempts from the AI research community to develop grounded interactive \emph{agents} that are capable of engaging in natural back-and-forth dialog with humans to assist them in completing real-world tasks~\citep{winograd1971procedures,narayan2017towards, levinson2019tom,chen2020ask}.

Importantly, the agent needs to understand when to initiate feedback requests if communication fails or instructions are not clear and requires learning new domain-specific vocabulary~\citep{Aliannejadi_convAI3,rao2018learning, narayan2019collaborative, jayannavar-etal-2020-learning}.
Despite all these efforts, the task is far from solved. 
For that reason, we propose the \name competition, which stands for \emph{Interactive Grounded Language Understanding in a collaborative environment}.

Specifically, the goal of our competition is to approach the following scientific challenge: 
\emph{How to build interactive agents that learn to solve a task while provided with grounded natural language instructions in a collaborative environment?}

By \textit{`interactive agent'} we mean that the agent is able to follow the instructions correctly, is able to ask for clarification when needed, and is able to quickly adapt newly acquired skills, just like humans are able to do while collaboratively interacting with each other.\footnote{An example of such grounded collaboration is presented in Figure~\ref{fig:A_B_dialog example}.} 

The described research challenge is naturally related, but not limited, to two fields of study that are highly relevant to the NeurIPS community: Natural Language Understanding and Generation (NLU / NLG) and Reinforcement Learning (RL).

\paragraph{Relevance of NLU/G} 
Natural language interfaces (NLIs) have been the ``holy grail'' of human-computer interaction and information search for decades~\citep{woods1972lunar, codd1974seven, hendrix1978developing}. The recent advances in language understanding capability~\citep{devlin2018bert, LiuRoberta_2019, clark2020electra, adiwardana2020towards, roller2020recipes, brown2020language}  powered by large-scale deep learning and increasing demand for new applications has led to a major resurgence of natural language interfaces in the form of virtual assistants, dialog systems,  semantic  parsing, and question answering systems~\citep{liu2017iterative, liu2018adversarial, dinan2020second, zhang2019dialogpt}. The horizon of NLIs has also been significantly  expanding  from databases~\citep{copestake1990natural} to,  knowledge  bases~\citep{berant2013semantic}, robots~\citep{tellex2011understanding}, Internet of Things (virtual  assistants like  Siri  and Alexa), Web service APIs~\citep{su2017building}, and other forms of interaction~\citep{fast2018iris, desai2016program, young2013pomdp}.
Recent efforts have also focused on interactivity and continuous learning to enable agents to interact with users to resolve the knowledge gap between them for better accuracy and transparency. This includes systems that can learn new task from instructions~\citep{li-etal-2020-interactive}, assess their uncertainty~\citep{yao-etal-2019-model}, ask clarifying questions~\citep{ Aliannejadi_convAI3, aliannejadi2021building} and seek and leverage feedback from humans to correct mistakes ~\citep{elgohary-etal-2020-speak}.

\paragraph{Relevance of RL} Recently developed RL methods celebrated successes for a number of tasks~\citep{bellemare2013arcade, mnih2015human, mnih2016asynchronous, silver2017mastering, hessel2018rainbow}. One of the aspects that helped to speed up RL methods development is game-based environments, which provide clear goals for an agent to achieve in flexible training settings. However, training RL agents that can follow human instructions has  attracted fewer exploration~\citep{chevalier2019babyai,cideron2019self,hu2019hierarchical, chen2020ask, shu2017hierarchical}, due to complexity of the task and lack of proper experimental environments.
~\citet{shu2017hierarchical} proposed a hierarchical policy modulated by a stochastic temporal grammar for efficient multi-task reinforcement learning where each learned task corresponds to a human language description in Minecraft environment. The BabyAI platform~\citep{chevalier2019babyai} aims to support investigations towards learning to perform language instructions with a simulated human in the loop. ~\citet{chen2020ask} demonstrated that using step-by-step human demonstrations in the form of natural language instructions and action trajectories can facilitate the decomposition of complex tasks in a crafting environment.

\paragraph{Minecraft as an Environment for Grounded Language Understanding}
\citet{szlam_why_2019} substantiated the advantages of building an open interactive assistant in the sandbox construction game of Minecraft instead of a ``real world'' assistant, which is inherently complex and inherently costly to develop and maintain. The Minecraft world's constraints (e.g., coarse 3-d voxel grid and simple physics) and the regularities in the head of the distribution of in-game tasks allow numerous scenarios for grounded NLU research~\citep{yao2020imitation, srinet-etal-2020-craftassist}. Minecraft is an appealing competition domain due to its popularity as a video game, of all games ever released, it has the second-most total copies sold. Moreover, since it is a popular game environment, we can expect players to enjoy interacting with the assistants as they are developed, yielding a rich resource for a human-in-the-loop studies.
Another important advantage of using Minecraft is the availability of the highly developed set of tools for logging agents interactions and deploying agents for evaluation with human-in-the-loop, including: 
\begin{itemize}[nosep]
    \item \textit{Malmo}~\citep{johnson2016malmo}: a powerful platform for AI experimentation;
    \item \textit{Craftassist}~\citep{gray_craftassist_2019}: a framework for dialog-enabled interactive agents;
    \item \textit{TaskWorldMod}~\citep{ogawa-etal-2020-gamification}: a platform for situated task-oriented dialog data collection using gamification; and
    \item \textit{MC-Saar-Instruct}~\citep{kohn2020mc}: a platform for Minecraft Instruction Giving Agents.
\end{itemize}

\noindent
Besides, mainly due to the success of previous competitions~\citep{guss2019minerlcomp,perez2019multi}, Minecraft is a widely used environment by the RL community for experimentation with (mainly single) agents trained by demonstration. Therefore, using Minecraft would set a low barrier for the RL community to contribute to \name. To simplify the competition settings and possibly lower the entry bar for the NLU/NLG community, we will use simulated Blocks World in Minecraft~\citep{jayannavar-etal-2020-learning}.

\paragraph{Relevance to Real Live Scenarios and Societal Impact}
Several important real-life scenarios have the potential to benefit from the results of our competition:
\begin{itemize}
    \item \textbf{Education:} 
    \textit{Minecraft: Education Edition}\footnote{\url{https://education.minecraft.net/}} is a game-based learning platform that promotes creativity, collaboration, and problem-solving in an immersive digital environment. As of 2021, educators in more than $115$ countries are using Minecraft across the curriculum. As stated in~\citet{url-minecraft-edu}, adding AI elements to this educational platform will move its potential to a new level. AI applications have the power to become a great equalizer in education. Students can get personalized education and scaffolding while being less dependent on uncontrollable factors such as the quality of their teachers or the amount of help they receive from their caregivers.

    \item \textbf{Robotics:} \citet{bisk-etal-2016-natural} proposed a protocol and framework for collecting data on human-robot interaction through natural language. The work demonstrated the potential for unrestricted contextually grounded communications between human and robots in blocks world. Developing robots to assist humans in different tasks at home has attracted much attention in the Robotics field~\citep{stuckler2012robocup}. In fact, the Robocup@Home\footnote{\url{https://athome.robocup.org/}} and the Room-Across-Room\footnote{\url{https://ai.google.com/research/rxr/habitat}} have run for several years.
    Given that the main human-robot interaction is through dialog, and the robot is supposed to assist the human in multiple tasks, we envision \name to enable more effective task grounded dialog training between human and robots.

\end{itemize}

\subsection{Novelty}

There is a long history of competitions focused on NLU/G tasks. Especially in recent years we have seen a large number of challenges dedicated to open-domain dialog systems~\citep{10.1145/3465272,scai-2020-international,spina2019cair,chuklin2018proceedings,arguello2018second}, such as ConvAI~\citep{burtsev2020conversational}, ConvAI2~\citep{dinan2020second}, ConvAI3: Clarifying Questions for Open-Domain Dialogue Systems (ClariQ)~\citep{Aliannejadi_convAI3,aliannejadi2021building}, as well as a series of competitions of the Alexa Prize\footnote{\url{https://developer.amazon.com/alexaprize}}. There are great efforts in the community to advance task-oriented dialogs by suggesting competitions, such as the Dialog System Technology Challenge (DSTC-8)~\citep{kim2019eighth}; benchmarks and experimental platforms, e.g., Convlab, which offers the annotated MultiWOZ dataset~\citep{budzianowski2018multiwoz} and associated pre-trained reference models~\citep{lee2019convlab}. There are fewer attempts to study multi-modal dialog systems, e.g., Situated Interactive Multi-Modal Conversational Data Collection And Evaluation Platform (SIMMC)~\citep{crook2019simmc} or Audio Visual Scene-Aware Dialog Track in DSTC8 \citep{hori2018audio}.

There are a number of RL competitions such as MineRL~\citep{guss2019minerlcomp} and MARLO~\citep{perez2019multi} that leverage the Minecraft environment. RL approaches have also been tried for text games environments, such as TextWold~\citep{yuan2019interactive}\footnote{\url{https://www.microsoft.com/en-us/research/project/textworld/}} and Learning in Interactive Games with Humans and Text(Light)~\citep{urbanek2019learning}\footnote{\url{https://parl.ai/projects/light/}}.

In comparison with previous efforts, to our knowledge, we are the first to propose a competition that tackles the task of grounded language understanding and interactive learning that brings together the NLU/G and RL research communities. The other key difference is our attempt to perform a human-in-the-loop evaluation as a final way for evaluating.

\subsection{Data}

\begin{figure}[t]
\centering
   \includegraphics[clip, width=1.0\columnwidth]{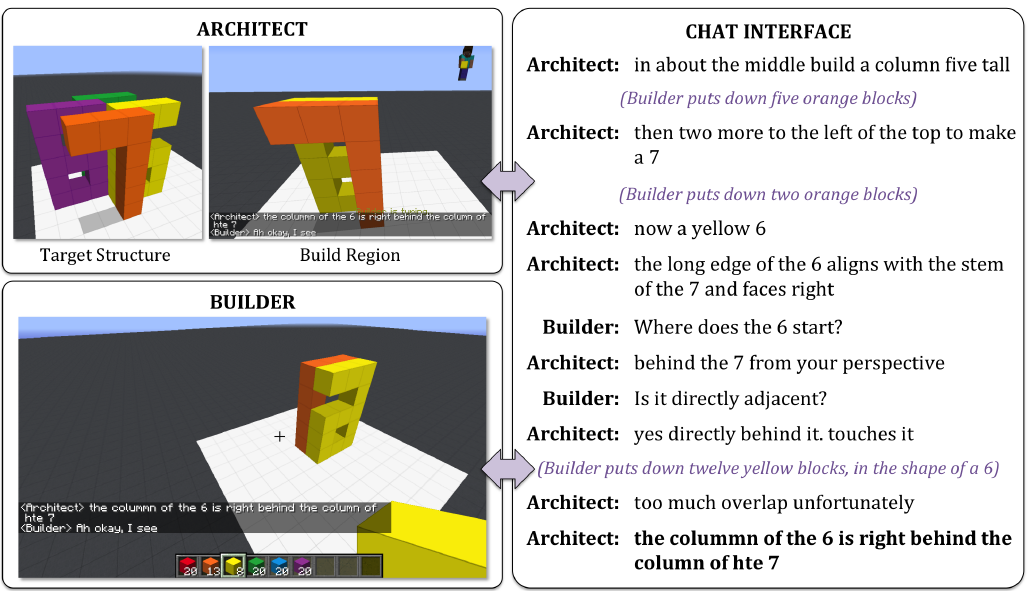}
   \caption{An example, taken from~\citep{narayan2019collaborative}, of the grounded collaborative dialog between Architect (A) and Builder (B), which is happening in 3D Block World.}
   \vspace{-0.5cm}
   \label{fig:A_B_dialog example}
\end{figure}

\paragraph{The general setup} \name is partially motivated by the HCRC Map Task Corpus~\citep{thompson_hcrc_1993}, which consists of route-following dialogs between an \emph{Instruction Giver} and a \emph{Follower}. These are given maps of an environment that differ on insignificant details. Namely, HCRC Map Task represents a cooperative asymmetric task that involves two participants. Further, we also use an \emph{Instruction Giver} and a \emph{Follower}, but call them \emph{Architect} and \emph{Builder} respectively, following the naming convention in~\citep{narayan2019collaborative}.

\citet{narayan2019collaborative} collected an openly available Minecraft dialog Corpus for a Collaborative Building Task\footnote{The dataset was released using Using Creative Commons Public Licenses \url{https://drive.google.com/drive/folders/16lDzswcQh8DR2jkQJdoVTK-RyVDFPHKa}. The authors are aware of the competition, and they are part of our Advisory Board.}. Figure~\ref{fig:A_B_dialog example} gives an example of the collected grounded dialogs. The authors used the following setup: the Architect is provided with a target structure that needs to be built by the Builder. The Architect provides instructions to the Builder on how to create the target structure and the Builder can ask clarifying questions to the Architect if an instruction is unclear~\cite{zhang-etal-2021-learning}. This dialog happens by means of a chat interface. The Architect is invisible to the Builder, but the Architect can see the actions of the Builder. The collected dataset has:
\begin{itemize} [nosep]
    \item $509$ collected human-to-human dialogs along with RGB observations, and inventory information;
    \item games played over the course of 3 weeks (approx. 62 hours overall) with $40$ volunteers. Each game took 8.55 minutes on average; and
    \item $163$ tasks for the Architect-Builder collaborative game.
\end{itemize}

\subsubsection{Data collection }
\label{sec:data_collection}

\paragraph{Extending the dataset with new tasks} We are working on creating new tasks for data collection that will be used to (1) extend the existing Minecraft Dialog Corpus~\citep{narayan2019collaborative} and (2) perform offline and online evaluation (as they are unseen by participants).
We provide up to $200$ newly designed target structures for building and each of them is associated with a difficulty level, ranging from \emph{Easy} to \emph{Normal} and \emph{Hard}. The difficulty level reflects the complexity of the target structure by considering the characteristics, such as the number and color diversity of the used blocks. This gives us a reasonable method to test the limit of our trained agents, including both architects and builders. Each target structure will be assigned to at least three different Architect-Builder pairs. Builders can manipulate the world within a $11 \times 11 \times 9$ sized build region and they have access to an inventory of $6$ colors of blocks and $20$ blocks of each color in each task.

\paragraph{Tooling for data collection} In order to enable data collection outside of lab settings, we extend the combination of tools proposed by~\cite{narayan2019collaborative} and~\cite{johnson2016malmo} for data logging. Specifically, we plan to provide an installer to the players to connect them through our server, which will enable them to play predefined tasks and store data afterwards.  Minecraft players will be invited through game forums and they will be rewarded with \textit{Minecoins}, the internal currency that is used in Minecraft. For collecting extended datasets at crowd-sourcing platforms, we will use the Craftassist tool~\citep{gray_craftassist_2019}, which can give us more data. However, potentially it can be of lower quality. In the case of extending the original dataset, we connect two humans. In the case of evaluation, we expose the trained agent to a single player.

Following the setup in \citet{narayan2019collaborative}, we record the progression of each task, corresponding to the construction of a target structure by an Architect and Builder pair, as a discrete sequence of game observations. Each observation contains the following information: 1) a time stamp, 2) the chat history up until that point in time, 3) the Builder's position (a tuple of real-valued $x$, $y$, $z$ coordinates as well as pitch and yaw angles, representing the orientation of their camera), 4) the Builder's block inventory, 5) the locations of the blocks in the build region, 6) screenshots taken from the Architect's and the Builder's perspectives. The detailed block locations and colors of each target structure will be linked to the corresponding building record.

\paragraph{Institutional Review Boards (IRB) Process}
As we plan to collect more data where human subjects are involved, we submitted our proposal for setting up a data collection pipeline to Institutional Review Boards to review our pipelines and approve the suggested consent form.

\noindent 
To summarize, we took all required steps to collect the final dataset before the beginning of the competition.

\subsection{Evaluation}
\label{sec:human-loop}
From an evaluation perspective, the interactive Architect-Builder collaboration to build a target structure is very convenient. Computing the Hamming distance between the built and the target structure can give us a straightforward measure of the success of the Architect-Builder collaboration~\citep{kiseleva2016predicting, kiseleva2016understanding}, which fits well to a large scale evaluation setup.
However, in this competition, we are particularly dedicated to bring a human in the loop to evaluate the trained agents' behavior.
To do that, we plan to pair a real player with the pre-trained agent to perform a building task. In addition to measuring the ability to achieve the target goal (building a provided structure), we will ask users to score the agent's performance.

\subsection{Tasks and application scenarios}

Given the current state of the field, our main research challenge (i.e., \emph{how to build interactive agents that learn to solve a task while provided with grounded natural language instructions in a collaborative environment}) might be too complex to suggest a reasonable end-to-end solution. Therefore, we split the problem into the following concrete research questions, which correspond to separate tasks that can be used to study each component individually before joining all of them into one system~\citep{jones1988look}:

\begin{enumerate}[label=\textbf{RQ\arabic*},nosep]
 \item \emph{How to teach}? \\ In other words, what is the best strategy for an Architect when instructing a Builder agent, such that the concept is reasonably explained? (The suggested task is presented in Section~\ref{sec:task1}).
 \item \emph{How to learn?} \\
  That is, what methods should be used to train a Builder that can follow given instructions from an Architect? \\
  This question can be further split into two sub-questions:
  \begin{enumerate}[label=\textbf{RQ2.\arabic*},nosep]
   \item \emph{How is a `silent' Builder able to learn?}\\
   A silent Builder follows instructions without the ability to ask for any clarification from the Architect. (The suggested task is presented in Section~\ref{sec:task2}).
   \item \emph{How is an `interactive' Builder able to learn?}\\
   An interactive Builder can ask clarifying questions to the Architect to gain more information about the task in case of uncertainty. (The suggested task is presented in Section~\ref{sec:task3}).
  \end{enumerate}
\end{enumerate}

Each of the suggested tasks is described below (Sections~\ref{sec:task1}, \ref{sec:task2}, \ref{sec:task3}) using the schema proposed by the NeurIPS template: (1) General Setup; (2)~Starting Code; (3)~Metrics for Evaluation; (4)~Baselines.
The participants are free to choose which task(s) they will solve. Each task will follow the pipeline presented in Figure~\ref{fig:flow} and will be evaluated separately.

\subsection{Task 1: Architect}
\label{sec:task1}

In this task, our goal is to develop an Architect that can generate appropriate step instructions based on the observations of environment and the Builder's behavior. 
At the beginning of each task, we give all the details of the target structure (e.g., types, colors and coordinated of blocks) to the Architect. The Architect needs to decompose the building process of this compound structure into a sequence of step instructions that the Builder can follow. During the interaction, the Architect has to compare the half-finished structure with the target structure and guide the Builder to complete the building of remaining components via generated instructions. The step instructions can be neither too detailed nor too general. In summary, the Architect is expected to be able to give instructions, correct the Builders’ mistakes and answer their questions by comparing the built structure against the target structure and by understanding the preceding dialog\footnote{We can see the similarity with newly published work on using human instructions to improve generalization in RL~\citep{chen2020ask}.}.

\subsubsection{Task setup}
We aim to generate a suitable Architect utterance, given access to 1) the detailed information of the target structure and 2) the entire game state context leading up to a certain point in a human-human game at which the human Architect spoke next. This task can be seen as a multimodal text generation, where the target structure, the built structure and the dialog history are input and the next Architect's utterance is the output. The model developed for this task can involve both language understanding and visual understanding depending on the methods for world state representations.

\subsubsection{Code}
The implementation of the baseline can be found using the following repository \url{https://github.com/prashant-jayan21/minecraft-dialogue-models}.

\subsubsection{Evaluation}
\paragraph{Automatic evaluation}
To evaluate how closely the generated utterances resemble the human utterances, we adopt standard BLEU scores~\citep{papineni2002bleu}. We also make use of the modified \textit{Precision} and \textit{Recall} of domain-specific keywords defined in~\citet{narayan2019collaborative}. The defined keywords are instrumental to task success, including colors, spatial relations, and other words that are highly indicative of dialog actions.

\paragraph{Human evaluation}
To better evaluate the quality of generated utterances, we will conduct human evaluations of Architect instructions. We will pair human participants with a trained architect model. The human builder will interact with the Architect and complete building tasks by following instructions from the Architect. After the target structure is successfully built or the interaction exceeds the maximum task time, the human builder will rate the experience from two different aspects: \textit{language naturalness} and \textit{instruction accuracy}. Each Architect model will be paired with three human builders and each Architect-human pair will be assigned three target structures to complete, corresponding to three different difficulty levels. For fair comparison, the human builders and target structures will be kept the same for all submitted Architect models. 

\subsubsection{Baseline}

\begin{figure}[t]
\centering
   \includegraphics[clip, width=0.6\columnwidth]{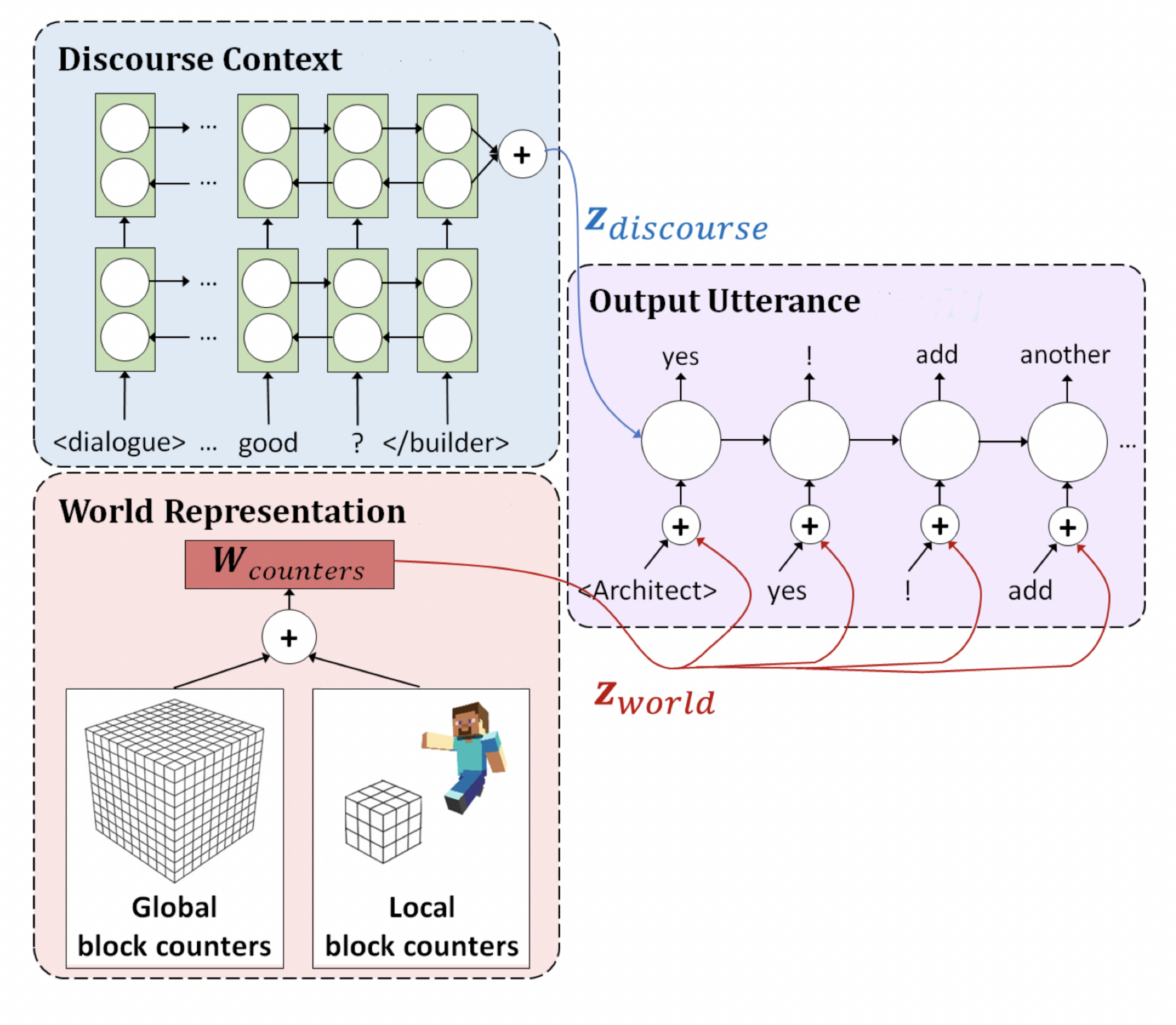}
   \caption{An overview of the full model combining global and local world representation variants from~\citep{narayan2019collaborative}.}
   \vspace{-0.5cm}
   \label{fig:architect-baseline}
\end{figure}

We provide two baselines that are presented in \citet{narayan2019collaborative}. 
The first baseline is a vanilla \textit{seq2seq} model that conditions the next utterance on the dialog context. A dialog history encoder is applied to obtain the context representation by compressing the dialog history consisting of multiple utterances. Speaker-speciﬁc start and end tokens ($\langle\text{A}\rangle$ and $\langle/\text{A}\rangle$ or $\langle\text{B}\rangle$ and $\langle/\text{B}\rangle$) are added to each dialog turn. The tokens are fed through a bidirectional \ac{RNN} to obtain the final representation. For the decoding step, a decoder \ac{RNN} is used to generate the next Architect utterance conditioned on the representation from the encoding step using beam search.

The second baseline is also a \textit{seq2seq} model, and it makes use of not only the text information but also the world state information at each step. The world state representations take into account the Hamming distance between the target structure and built structure and it also tells how the target structure can be constructed successfully given the current state of the built structure, e.g., which blocks needs to be placed or removed. \citet{narayan2019collaborative} consider two variants of block counters that capture the current state of the built structure: \emph{Global block counters} and \emph{Local block counters}. Global block counters are computed over the whole build region ($11 \times 11 \times 9$) while local block counters encode spatial information of a cube of blocks ($3 \times 3 \times 3$) first and concatenate the cube representations to get the final world state representation. The world state representations are concatenated to the word embedding vector that is fed to the decoder at each decoding step as shown in Figure~\ref{fig:architect-baseline}.

\paragraph{Preliminary results}

We report the results on the test set presented in~\citep{narayan2019collaborative} in Table~\ref{Table:architect_bleu}. The hyper-parameters of architect models have been fine-tuned on the validation set. By augmenting both the global world state and local world state, Seq2Seq with global and local information managed to show noticeable improvements on each of the automatic metrics. The provided baseline definitely leaves room for improvement. All the architect models will be re-evaluated after we collect a larger dialog corpus.

 \begin{table}[ht!]
  \centering
  \resizebox{0.9\linewidth}{!}{
  \begin{tabular}{ l c*{9}{c}}
    \toprule
 & \multicolumn{4}{c}{BLEU} & \multicolumn{4}{c}{Precision/ Recall}  \\
  \cmidrule(r){2-5}
  \cmidrule(r){6-9}
 \makecell{Metrics}  &\makecell{BLEU-1} & \makecell{BLEU-2} & \makecell{BLEU-3} & \makecell{BLEU-4} & \makecell{all keywords} &\makecell{colors} &\makecell{spatial}  &\makecell{dialog}\\
\midrule
Seq2Seq    &15.3  &7.8     &4.5    &2.8  &11.8/11.1   &8.1/17.0 &9.3/8.6  &17.9/19.3  \\
\makecell{+global \& local} &15.7  &8.1     &4.8    &2.9  &13.5/14.4   &14.9/28.7  &8.7/8.7  &18.5/19.9  \\
   \bottomrule
  \end{tabular}
  }
\caption{BLEU and term-specific precision and recall scores on the test set, originally reported in \citep{narayan2019collaborative}, which were able to reproduce.} 
\label{Table:architect_bleu}
\end{table}

\subsection{Task 2: Silent Builder}
\label{sec:task2}

\subsubsection{Setup}
\begin{figure}[t]
\centering
   \includegraphics[clip, width=0.95\columnwidth]{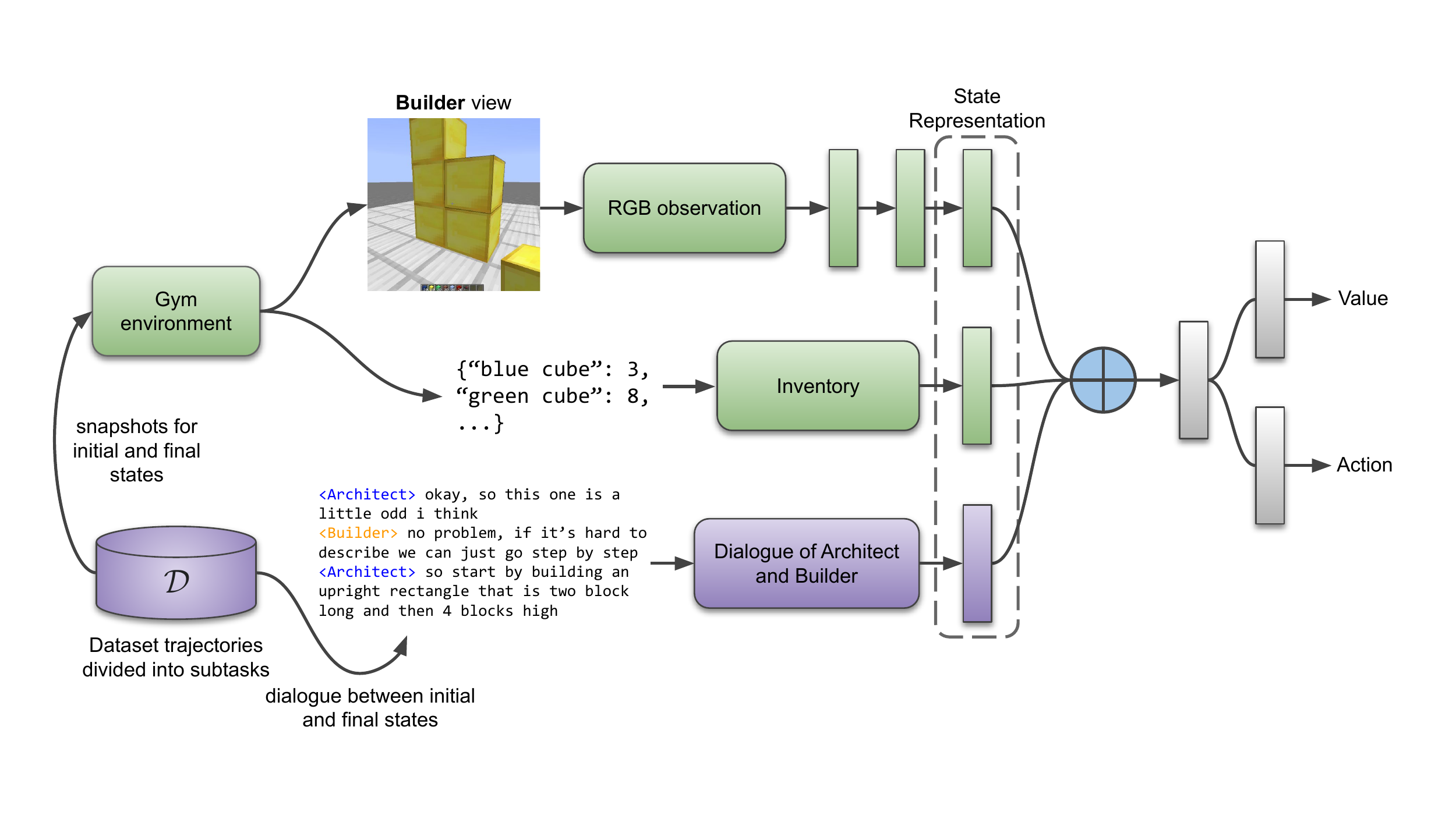}
   \caption{The overall pipeline for suggested baselines for silent builder}
   \label{fig:silent-builder}
\end{figure}

The overall setup for training initial baselines for the silent Builder that will be used for comparison is presented in Figure~\ref{fig:silent-builder}.

\subsubsection{Interface/Starting Code}
\label{sec:code_rl}
We provide a set of Gym environments\footnote{\url{https://blog.openai.com/universe/}}, which are based on sub-tasks extracted from the~\cite{narayan2019collaborative} dataset. An example of such sub-tasks can be building an L-like structure or building several red blocks, forming a heart. 
The state-space in which the agent operates consists of the state of the environment (e.g., RGB image of POV), the inventory and the annotated sub-goal, defined by the Builder-Architect dialog from the dataset.

In the current version of the environment, the agent has access to:
POV image $(64, 64, 3)$, inventory item counts $(6,)$, building zone occupation block types $(11, 11, 9)$, full chat embedding $(1024,)$ and agent $(X, Y, Z)$ position with two angles $(5,)$. A grid specifies the building zone. In each cell, we have an ID that defines the block type in that position (e.g., $0$ for air, $1$ for gold block, etc.). 

\begin{figure}
\begin{subfigure}{.25\textwidth}
  \centering
  \includegraphics[width=0.95\linewidth]{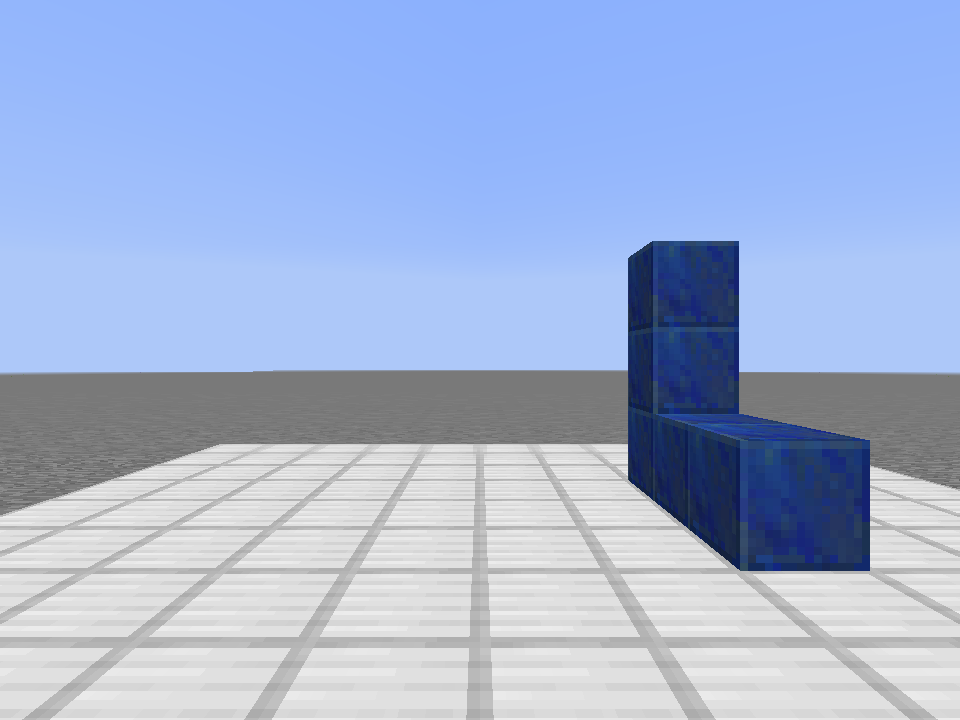}
  \caption{5 block L}
  \label{fig_env:1}
\end{subfigure}%
\begin{subfigure}{.25\textwidth}
  \centering
  \includegraphics[width=0.95\linewidth]{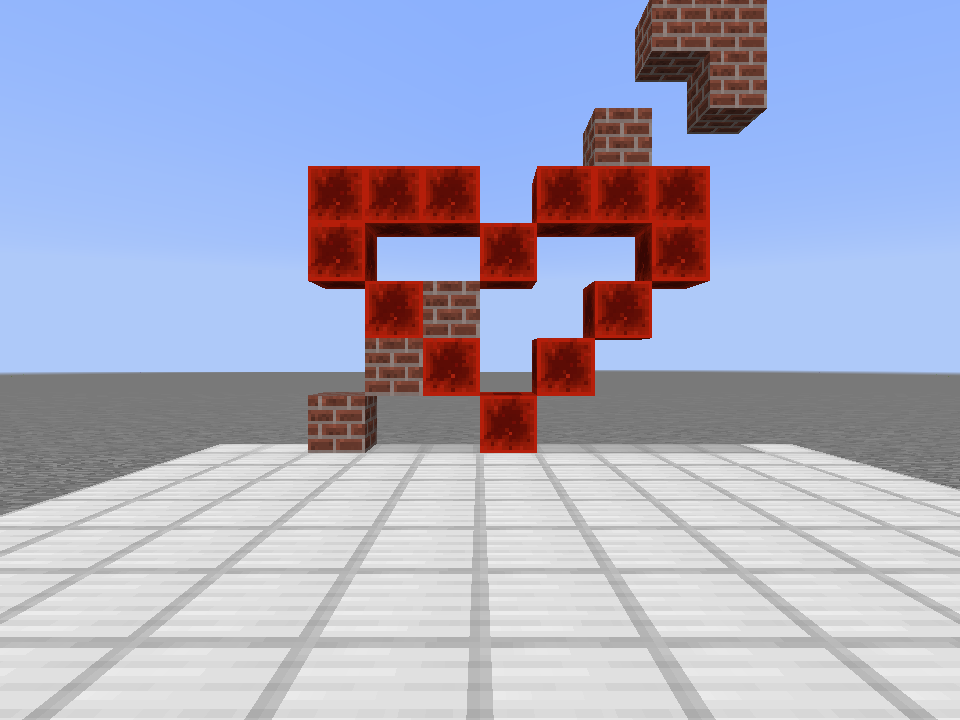}
  \caption{A ``broken heart''}
  \label{fig_env:2}
\end{subfigure}
\begin{subfigure}{.25\textwidth}
  \centering
  \includegraphics[width=0.95\linewidth]{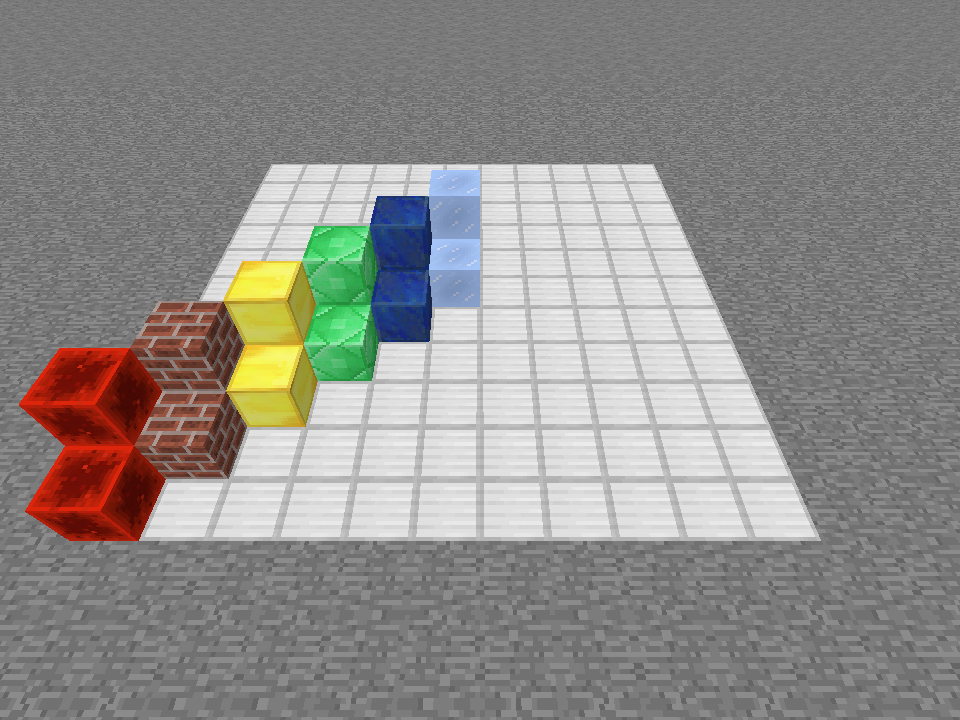}
  \caption{diagonal 3 block Ls}
  \label{fig_env:3}
\end{subfigure}%
\begin{subfigure}{.25\textwidth}
  \centering
  \includegraphics[width=0.95\linewidth]{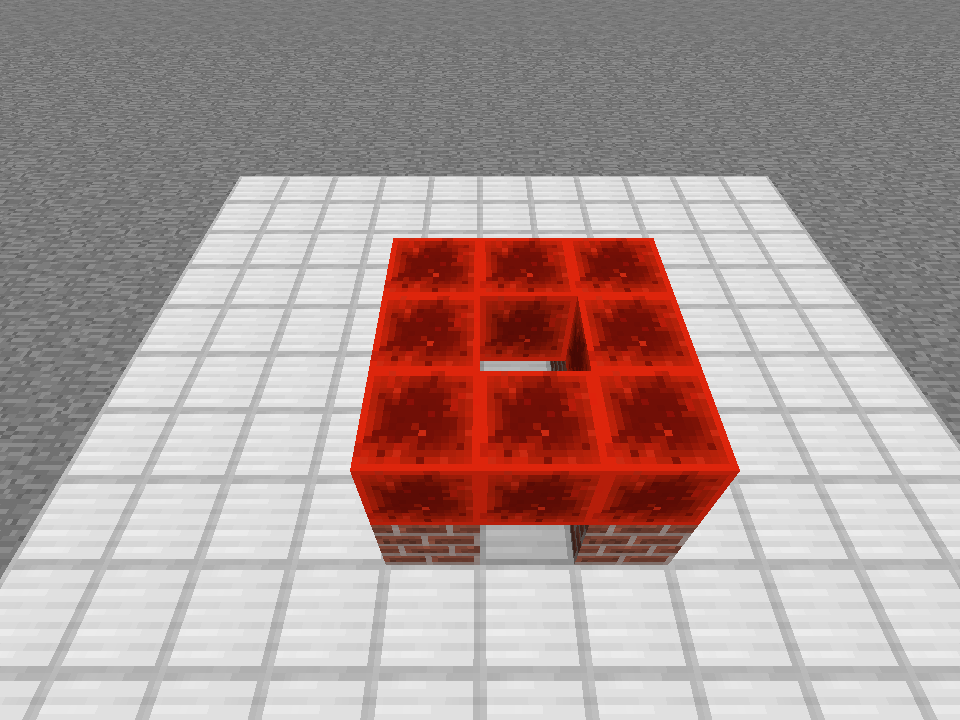}
  \caption{A table}
  \label{fig_env:4}
\end{subfigure}
\caption{Examples of sub-tasks in the environment}
\label{fig:env}
\end{figure}

The agent's action space might consist of all possible actions in Minecraft. For the current Builder baseline, we offer a discretized version of 18 actions:
\textit{noop, step forward, step backward, step right, step left, turn up, turn down, turn left, turn right, jump, attack, place block, choose block type 1-6}. We parametarize the environment by a sub-task ID which is extracted from the dataset and remap its custom block types with default Minecraft block types (e.g., gold block instead of yellow block, Redstone block instead of red ones, etc.)\footnote{A representative example of the environment with random agent behavior: \url{https://youtu.be/hqQ0ubbULWQ}}. In the future, we will adapt our environment to be visually consistent with the ~\citep{narayan2019collaborative} dataset to enable data-driven RL training. In Figure~\ref{fig:env} four sub-tasks for the RL agent are shown. 

For simplicity of the task, we allow an agent to build a target structure anywhere inside the building zone by comparing all relative translations and rotations of two zones. The environment calculates intersections over two zones and takes the maximum over these intersections. The agent then receives a reward according to the best match if it has been changed (i.e., non-zero reward if and only if the max match was changed since the last action). If the agent gets closer to the target structure in terms of the maximal match, it receives a reward $+2$. If the structure built by the agent moves further away in terms of a maximal match from the target one (e.g., the agent removes a correctly placed block), the agent gets a reward of $-2$. Otherwise, if the agent places/removes a block outside of a partially built structure (i.e., without changing maximal intersections of structures), it receives a reward of $-1$/$+1$ respectively. If the agent moves outside of the building zone, then the environment terminates immediately.


\paragraph{Architect simulator for training the silent Builder}
The correct instruction sequence to achieve the final goal is known, since the target structures are associated with dialogs. We annotated each dialog's sub-goals and stored them in a queue, where each sub-goal corresponds to one specific step instruction from the Architect. At each step, we pop up a sub-task (e.g., in about the middle, build a column five tall) and wait until the agent completes it. If the agent completes this sub-task, we pop up the next sub-task. We trained a matching model to decide if the current sub-task has been finished. 

\subsubsection{Evaluation}
\label{sec:eval_rl}

\paragraph{Automatic evaluation} For each subtask of the dataset, the submitted agent will be evaluated on the environment, with fixed initial and target states. We suggest the following three metrics by which the solution will be evaluated: 
\begin{itemize}
    \item The reward score $S_r$ is based on the average reward received by the agent in evaluation episodes, which is calculated as follows:

        \begin{equation}
            S_{r}=\frac{1}{N} \sum_{i=1}^{N}g_i,
        \end{equation}
        where N is a number of evaluation episodes, $g_i$ is episode reward, defined by
        
        \begin{equation*}
            g_i=\sum_{t=1}^{T}r_t.
        \end{equation*}
    
    \item The success rate score $S_s$ indicates the number completely solved subtasks:
    
        \begin{equation}
            S_{s}=\frac{1}{N}\sum_{i=1}^{N}c_i, 
        \end{equation}
        
        where 
        \begin{equation*}
        c_i = 
         \begin{cases}
           +1, &\text{if success,} \\
           0, &\text{otherwise.}
         \end{cases}
        \end{equation*}
    
    \item Completion rate score $S_{c}$:

        \begin{equation}
            S_{c}=\frac{1}{N}\sum_{i=1}^{N}1-\rho_i, 
        \end{equation}
        where $\rho$ is a normalized Hamming distance between target and built structures.
\end{itemize}

\subsubsection{Baselines and Results}
\label{sec:rl_baseline}
The described baselines and preliminary results are obtained for the settings of the silent Builder.

\paragraph{Preliminary Results}
With the environment described above, we conducted baseline experiments by setting one goal per trained agent. Specifically, we started from a straightforward case where the goal was to build a simple L-shaped structure of 5 blocks. Since the task was fixed per agent, we did not expect the agent to extract any meaningful information from the conversation embedding. Therefore, for the first baseline, we excluded this information from the agent's observation. 

\begin{figure}[t]
\centering
   \includegraphics[clip, width=0.6\columnwidth]{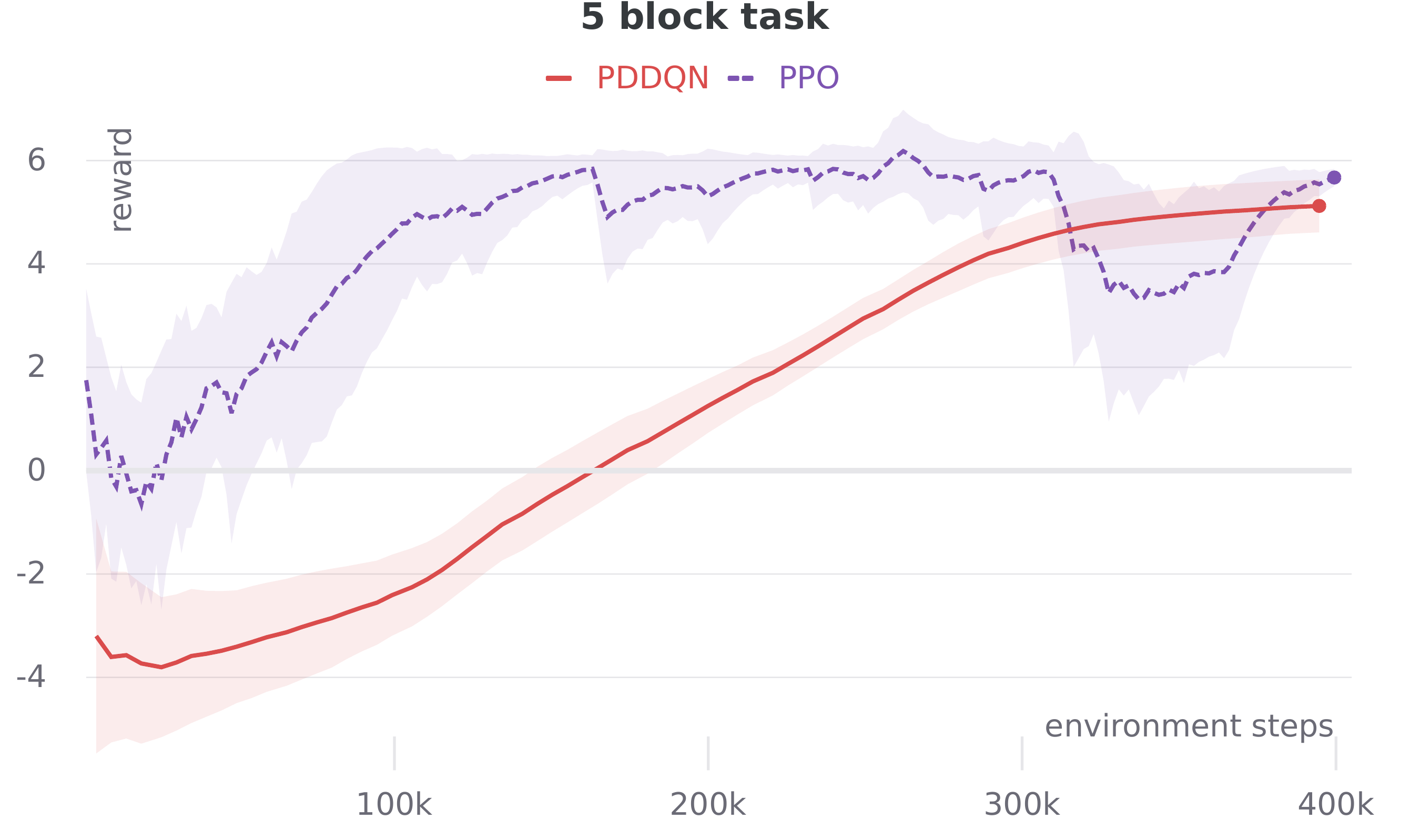}
   \caption{Performance of the PDDQN and PPO agents on 5 block task (reward averaged over 3 runs).}
   \vspace{-0.5cm}
   \label{fig:pddqn}
\end{figure}

As baselines, we trained Proximal Policy Optimization~(PPO)~(\cite{ppo}) and a variant of Rainbow~(\cite{rainbow_dqn}), PDDQN with experience replay prioritization, dueling architecture, and multi-step return enabled. We trained an agent for 400k environment steps since the task is relatively simple. For the experiments, we used the RLlib framework (\citep{rllib}) as it provides a solid set of high-performance RL implementations. The performance plot of the agent is shown in Figure~\ref{fig:pddqn}.

\paragraph{Planned Baselines}
We will provide baselines trained on a full set of sub-tasks using Rainbow and PPO algorithms, and their pre-trained on human demonstration versions DQfD~(\citep{dqfd}), ForgER~(\cite{forger}) and POfD~(\citep{pofd}).  The baselines will be implemented based on the silent builder scheme. 

\subsection{Task 3: Interactive Builder}
\label{sec:task3}

\begin{figure}[t]
\centering
   \includegraphics[clip, width=0.95\columnwidth]{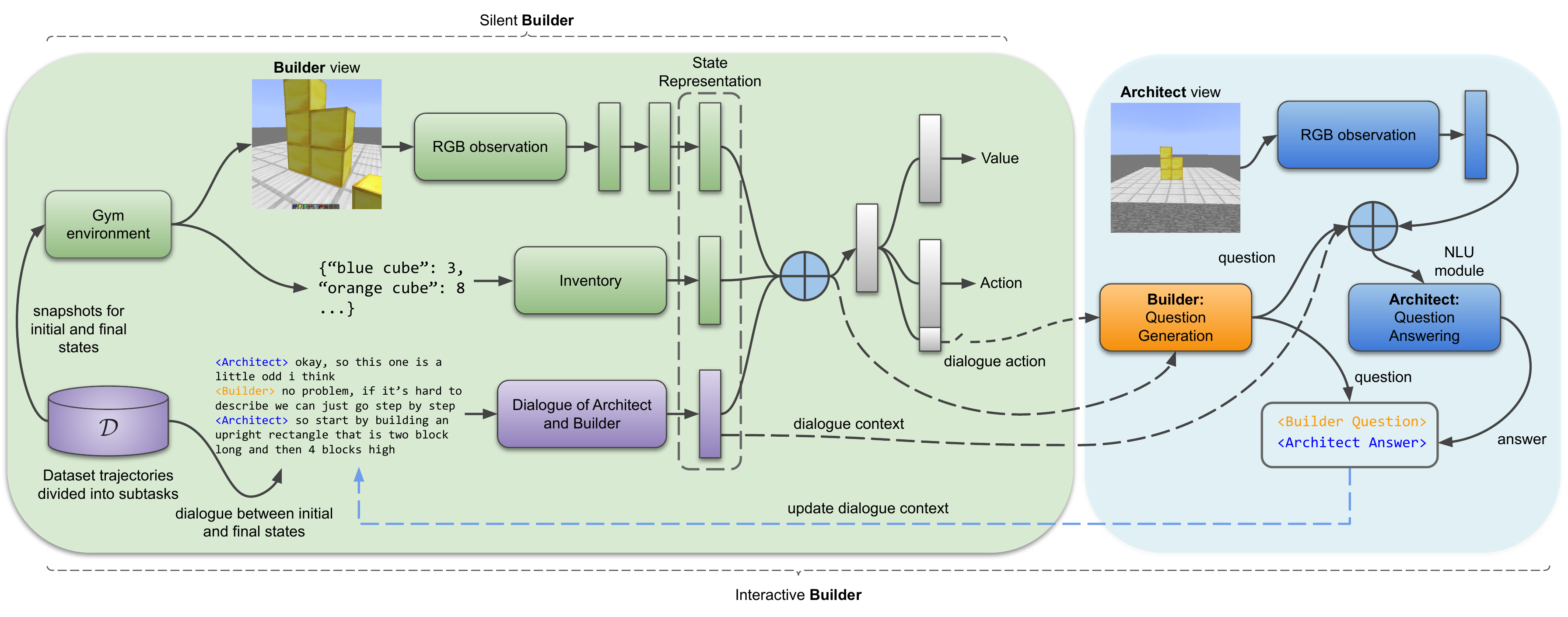}
   \caption{The overall pipeline for interactive builder. Builder(QG) and Architect(QA) are trained offline and their weights are frozen.}
   \vspace{-0.5cm}
   \label{fig:interactive-builder}
\end{figure}

The problem of developing an interactive Builder can be viewed as a combination of an extended Architect and an upgraded silent Builder.
Figure~\ref{fig:interactive-builder} shows our view for the solution, which is a work in progress that we plan to finish before the start date of the competition and use as an internal baseline.
However, we do not intend to share the presented suggested solution with the participants to avoid biasing their thinking and creativity in approaching the problem. We also understand that this task is complex, hence it might not attract as much attention as the previous two.

In terms of starting code, the Gym environment is described in Section~\ref{sec:code_rl}, and the same evaluation metrics for automatic evaluation can be used as presented in Section~\ref{sec:eval_rl}.

\subsection{Tutorial and documentation}

Our competition page \url{future_page} will contain detailed instructions on: (1)~provided datasets; (2)~the developed Gym environment; (3)~baselines; (4) leader-board and other required resources.

\section{Organizational aspects}
\label{sec:organization}
\subsection{Protocol}

\subsubsection{Submission}
To orchestrate the submissions' evaluation, we will use one of the open-source platforms for organizing machine learning competitions. Currently, we are exploring options with the following platforms: \url{AICrowd.com}, \url{Kaggle.com}, and \url{Codalab.org}. While making our final decision about the platform for \name, we will consider the following important aspects:
\begin{itemize}[nosep]
    \item Participants should share their source code (preferably a git repository to see versioning as well) and pre-trained models throughout the competition with the organizers;
    \item Automated way to trigger our offline evaluation pipeline;  
    \item Easy way to package the software runtime for the suggested solutions (preferably through docker); and
    \item The ability to reliably orchestrate user submissions.
\end{itemize}
\noindent
We will make our final decision about the platform after proposal acceptance.

\subsection{Competition Pipeline}

\begin{figure}
\centering
   \includegraphics[clip, width=1.0\columnwidth]{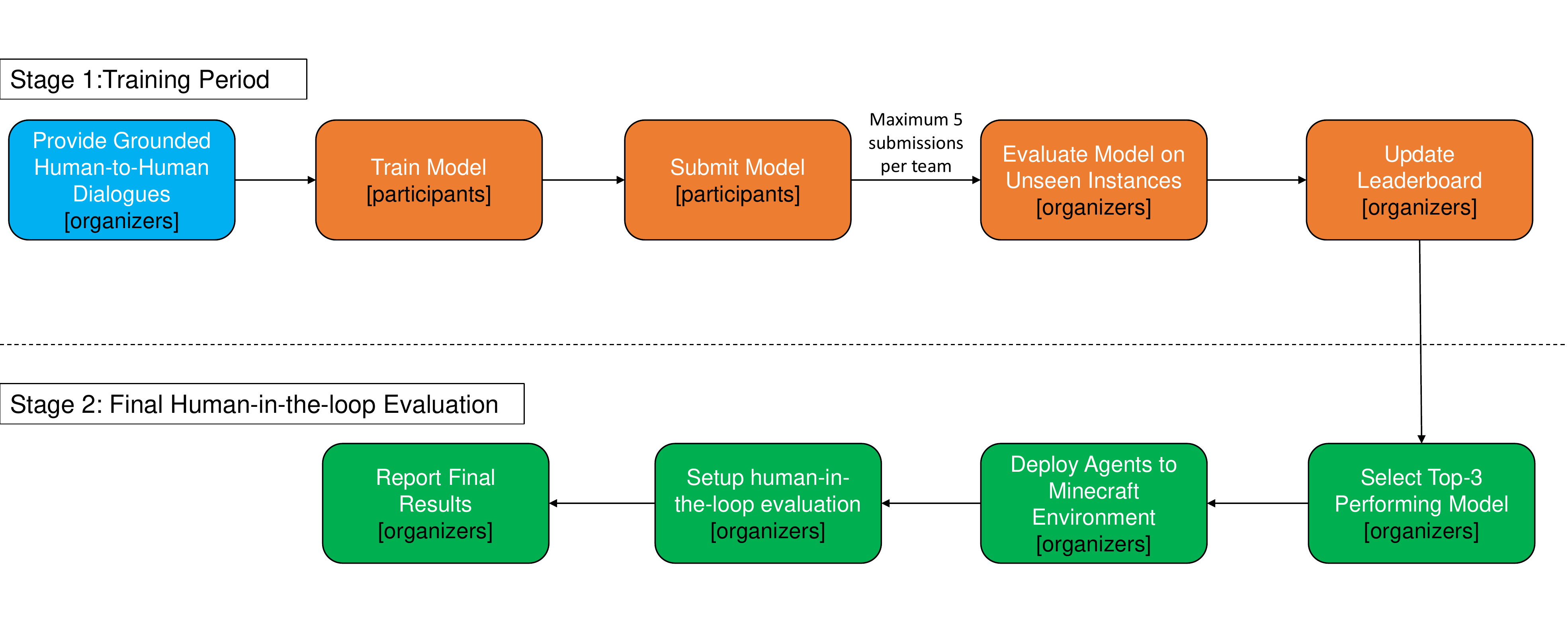}
   \caption{The general flow of the \name competition which consists of two main stages: (1)~Training period; (2)~Final human-in-the-loop evaluation.}
   \vspace{-0.5cm}
   \label{fig:flow}
\end{figure}

Figure~\ref{fig:flow} details the general structure of the \name competition, which can be split into two main Stages.
\begin{itemize}[nosep]
    \item Stage 1: Training Period, which includes the following main actions.
       \begin{itemize}[nosep]
           \item \emph{`Provide Grounded Human-to-Human Dialogs'}: Participants need to register for the competition on our website, where they indicate which task(s) they are planning to solve. Next, we will send them the collected Human-to-Human dialogs. They can find the code for the available baselines on our website, as well as documentation on how to use the supplied Gym environment.
           \item \emph{`Train Model'}: This part is done by the participants (they may request compute resources from Azure, which we provide).
           \item \emph{`Submit Model'}: Participants should decide which models they want to test and then submit them (with a maximum number of submissions per task per team of $5$).
           \item \emph{`Evaluate Model on Unseen Instances'}: The organizers do this part. As a result of this action, the participants will receive the evaluation score for their method, and the organizers will update the leader board.\\
           As a result of the training period, we will pick the top-3 performing solutions per task for the next stage.
       \end{itemize}
    \item Stage 2: Final Human-in-the-loop evaluation, which is fully performed by the organizers and includes the following actions\footnote{This stage might happen after the NeurIPS to ensure that participants have enough time to develop their solutions.}.
           \begin{itemize}[nosep]
           \item The organizers will deploy the selected solutions into the Minecraft environment (using the CraftAssist toolkit).
           \item The organizers will set up the task and will invite players to interact with the developed Architect(s) and Builders(s).
           \item The organizers will report the final results based on the score provided by humans (the final scoring schema for the human evaluation is under development, and it varies per task).
           \end{itemize}
\end{itemize}
The winners will be announced after the results of human-in-the-loop experiments are collected for each of the tasks.

\subsection{Rules}
The competition aims to develop agents that can converse with humans to collaborate on a common goal successfully.  Furthermore, to ensure that the competition maximally benefits the research community, we plan to open-source the final solutions to enable future reproducibility. We introduce the following rules in an attempt to capture the spirit of the competition, and any submissions found to be violating these rules may be deemed ineligible for participation by the organizers.
\begin{itemize}[nosep]
    \item Participants may re-use open-source libraries or models with proper attribution.  Conversely, contestants will open source any new models or libraries developed for the competition.
    \item Contestants are not limited to the provided dataset, but
    \begin{enumerate}[nosep]
        \item Participants should clearly describe all data; and
        \item Contestants must release (open-source) all non-public additional data used to train solution with their submissions, or any public data which has been aggregated or processed.
    \end{enumerate}
\end{itemize}

\paragraph{}
\noindent \textbf{To prevent cheating} we introduce the following constraints:
\begin{itemize}[nosep]
    \item The dataset of `Unseen Instances' used for offline evaluation is never shared with the participants;
    \item The participants can submit their solutions for offline evaluation at most $5$ times to prevent any reverse engineering of the results;
    \item The final evaluation is done with a human-in-the-loop setup, where at least 5 humans are playing with the submitted agents over double-blind trials to ensure fairness; and
    \item In suspicious cases, we will look over the submitted code.
\end{itemize}

subsection{Readiness}
The following important milestones are already taken care of at the moment of the proposal submission:
\begin{itemize}[nosep]
    \item The data collection tools have been implemented (in collaboration with Facebook AI Research and Microsoft Research), which are going to be used to extend the current dataset~\citep{narayan2019collaborative} at Stage 1 and for the human-in-the-loop-evaluation at Stage 2;
    \item The initial Gym environment for training the Builder has been implemented and tested;
    \item The IRB process has been started to ensure proper data collection which involves humans;
    \item The initial baselines for the Architect and the silent Builder have been released;
    \item The work on increasing the number of tasks with different difficulty levels has been started;
    \item Organizers started looking into a suitable platform to orchestrate the submissions and evaluation.
\end{itemize}
In case the proposal is accepted to the NeurIPS competition track, we do not envision roadblocks preventing the competition's execution.

\subsection{Competition promotion}

\paragraph{Main mailing lists}
We plan to distribute the call to general technical mailing lists, company internal mailing lists, and institutional mailing lists to promote participation in the competition. We will directly contact leaders of the relevant groups in universities and industry labs to promote the competition.

\paragraph{Attracting Participants from Underrepresented Groups} We plan to partner with a number of affinity groups to promote the participation of groups that are traditionally underrepresented at NeurIPS. Specifically, we will reach out to the following groups: 
\begin{itemize}[nosep]
    \item Black in AI (BAI);
    \item LatinX in AI (LXAI);
    \item Queer in AI;
    \item Women in Machine Learning (WiML); and
    \item Women and Underrepresented Minorities in Natural Language Processing (WiNLP);
\end{itemize}
We plan to reach out to other organizations, such as Data Science Africa, to work with them to increase the participation of individuals underrepresented in North American conferences and competitions.

\paragraph{Media} To increase general interest and excitement surrounding the competition, we will work with the media coordinators at Microsoft, Minecraft, Facebook, and the University of Amsterdam. Moreover, we will use personal media accounts to promote the \name competition.

\paragraph{Presentation at different venues} Organizational and advisory board members will include the information about the \name competition in their upcoming presentations.

\section{Resources}
\label{sec:resources}

\subsection{Organizing team}

As we noted above, one of the goals of the competition is to bring researchers from different sub-fields (NLU/G and RL) to work together. To ensure that our competition is interesting for researchers from all of these directions, the organizational team consists of researchers from different disciplines, such as NLP and RL. Moreover, the organizational team is distributed across different continents and countries, including the Netherlands, Russia, Great-Britain, and the United States. On top of that, our team is well balanced with respect to the following aspects:
\begin{itemize}[nosep]
    \item a great mix of academic and industrial researchers;
    \item a good mix of researchers at different stages of their career, including Ph.D. students, Post Docs and Senior Researchers.
\end{itemize}
Many of our team members have successfully organized previous challenges, including ConvAI1, ConvAI2, ClariQ (ConvAI3), DSTC7, DSTC8, MineRL, and Marlo. For some team members this is their first experience with organizing a challenge -- more experienced colleagues will support them. Moreover, the team led by Aleksandr Panov and Alexey Skrynnik (see Section~\ref{sec:organizers}), won the MineRL competition. Their experience can be used to improve \name and make it friendly for competitors.

\subsection{Organizers}
\label{sec:organizers}

Our team can be split into two main groups based on expertise: NLU/G and RL researchers.

\paragraph{\emph{The RL sub-team includes the following members:}}

\noindent
\paragraph{Alexey Skrynnik} is a Research Associate and Ph.D. student advised by Aleksandr Panov at Artificial Intelligence Research Institute FRC CSC of Russian Academy of Sciences. His current research focused on hierarchical reinforcement learning. Alexey previously won MineRL Diamond Competition 2019 as a leader of the CDS team. He leads the RL baselines development and is planning to make it a part of his dissertation, extending RL with NLP.

\noindent
\paragraph{Artem Zholus} is a Research Engineer and Master's student at the Cognitive Dynamic Systems laboratory at Moscow Institute of Physics and Technology. His research focuses on model-based reinforcement learning with application to robotics. Artem has worked on many machine learning projects, such as Reinforcement Learning for drug discovery and generative models. In this project, he has developed the environment for RL agents and trained several RL algorithms as baselines as part of the CDS team.

\noindent
\paragraph{Aleksandr Panov} is head of the Cognitive Dynamic Systems Laboratory at Moscow Institute of Physics and Technology. Aleksandr managed the CDS team that won the MineRL Diamond Competition 2019. Currently, Aleksandr works in model-based RL and the use of RL methods in the task of navigation in indoor environments.

\noindent
\paragraph{Katja Hofmann} is a Principle Researcher at the Machine Intelligence and Perception group at Microsoft Research Cambridge. Her research focuses on reinforcement learning with applications in video games, as she believes that games will drive a transformation of how people interact with AI technology. She is the research lead of Project Malmo, which uses the popular game Minecraft as an experimentation platform for developing intelligent technology, and has previously co-organized two competitions based on the Malmo platform. Her long-term goal is to develop AI systems that learn to collaborate with people, to empower their users and help solve complex real-world problems.

\noindent
\paragraph{\emph{The NLU/G sub-team consists of the following members:}}

\noindent
\paragraph{Ziming Li} is a Postdoc at University of Amsterdam. His main research interest is developing advanced dialogue systems, including dialogue policy learning and evaluation. He is also interested in the fields of conversational search and reinforcement learning.
\noindent
\paragraph{Julia Kiseleva} is a Senior Researcher in Microsoft Research. Her research interests are in natural language processing, information retrieval, and machine learning, with a strong focus on continuous learning from user feedback and interactions. Julia has been involved in many initiatives such as a serious of Search-Oriented Conversational AI (SCAI) workshop, which was continuously organized from 2015 - 2020. She also co-organized the recent challenge ConvAI3: Clarifying Questions for Open-Domain Dialogue Systems (ClariQ) at EMNLP2020.
\noindent
\paragraph{Maartje ter Hoeve} is a PhD candidate at the University of Amsterdam. Her main research interest is how we can learn from humans and their cognition to improve our NLP and IR systems. She has a background in both Linguistics and Artificial Intelligence.

\noindent
\paragraph{Mikhail Burtsev} is a Head of Neural Networks \& Deep Learning lab at MIPT, Moscow. His current research interests: application of neural nets and reinforcement learning in the NLP domain. He is a faculty advisor of MIPT team participating in Alexa Prize Challenges 3 and 4.  He proposed and organised Conversational AI Challenges: ConvAI 1 (NIPS 2017), ConvAI 2 (NeurIPS 2018), ConvAI 3 (EMNLP 2020). 

\noindent
\paragraph{Mohammad Aliannejadi} is a post-doctoral researcher at the University of Amsterdam (The Netherlands). His research interests include single- and mixed-initiative conversational information access and recommender systems. Previously, he completed his Ph.D.~at Universit\`a della Svizzera italiana (Switzerland), where he worked on novel approaches of information access in conversations. He co-organized the Conversational AI Challenge, ConvAI 3 (EMNLP 2020).

\noindent
\paragraph{Shrestha Mohanty} is a Machine Learning Engineer at Microsoft. She primarily works in areas of machine learning, deep learning, and natural language processing, including topics such as personalization, dialogue systems and multilingual language models. She is also interested in and has worked on problems at the intersection of machine learning and healthcare. Prior to Microsoft, she completed her master’s in information science from University of California, Berkeley.

\noindent
\paragraph{Arthur Szlam} is a Research Scientist at Facebook AI Research.  He works on connecting perception, memory, language, and action in artificial agents.  Prior to joining Facebook, he was on the faculty of the City College of New York (CCNY), and was the recipient of a Sloan research fellowship. He has been a co-organizer of previous ConvAI challenges.

\noindent
\paragraph{Kavya Srinet} is a Research Engineering Manager at Facebook AI Research working towards a long-term goal of developing interactive assistants. She works on building assistants that can learn from interactions with humans. Prior to FAIR, she was a Machine Learning Engineer at the AI Lab at Baidu Research, where she worked on speech and NLP problems. Kavya was at the Allen Institute for Artificial Intelligence for a summer before that working on learning to rank for Semantic Scholar. Kavya did her graduate school from Language Technology Institute at Carnegie Mellon University, where she worked on areas of machine translation, question answering and learning to rank for graphs and knowledge bases.

\noindent
\paragraph{Yuxuan Sun} is a Research Engineer in Facebook AI Research (FAIR). His research interests lie in natural language processing, neural symbolic learning and reasoning, and human-in-the-loop learning for embodied agents.

\noindent\paragraph{Michel Galley} is a Principal Researcher at Microsoft Research. His research interests are in the areas of natural language processing and machine learning, with a focus on conversational AI, neural generation, statistical machine translation, and summarization. He co-authored more than 70 scientific papers, many of which appeared at top NLP, AI, and ML conferences. His organizational experience includes service as sponsorship chair at EMNLP-16, workshop chair at NAACL-21, challenge co-organizer at the Dialog System Technology Challenge (DSTC7) workshop at AAAI-19, and workshop chair for DSTC8 (at AAAI-20). He has been part of the DSTC challenge steering committee since 2020. He also served as area chair at top NLP conferences (ACL, NAACL, EMNLP, CoNLL), as action editor for the TACL journal (2020-), and will serve as area chair at NeurIPS-21.

\noindent
\paragraph{Ahmed Awadallah} is a Senior Principal Research Manager at Microsoft Research where he leads the Language \& Information Technologies Group. His research has sought to understand how people interact with information and to enable machines to understand and communicate in natural language (NL) and assist with task completion. More recently, his research has focused learning form limited annotated data (e.g., few-shot learning  and transfer learning) and from user interactions (e.g. interactive semantic parsing) . Ahmed’s contributions to NLP and IR have recently been recognized with the 2020 Karen Spärck Jones Award from the British Computer Society. Ahmed regularly serves as (senior) committee, area chair, guest editor and editorial board member at many major NLP and IR conferences and journal.

\subsection{Advisory Board}

\paragraph{Julia Hockenmaier} is an associate professor at the University of Illinois at Urbana-Champaign. She has received a CAREER award for her work on CCG-based grammar induction and an IJCAI-JAIR Best Paper Prize for her work on image description. She has served as member and chair of the NAACL board, president of SIGNLL, and as program chair of CoNLL 2013 and EMNLP 2018.

\noindent
\paragraph{Bill Dolan} is Partner Researcher Manager at Microsoft Research, where he manages the Natural Language Processing group. He has worked on a wide variety of problems, including the acquisition of structured common-sense knowledge from free text, paraphrasing, text rewriting to improve grammar and style, and most recently on data-driven, grounded approaches to handling dialog. He has helped organize a number of research community events over the years, including the RTE challenges and the ``Understanding Situated Language in Everyday Life" summer research institute with the University of Washington, as well as running the Microsoft Research graduate fellowship program from 2014-2017.

\noindent
\paragraph{Ryen W. White} is a Partner Research Area Manager at Microsoft Research, where he leads the Language and Intelligent Assistance research area. He led the applied science organization at Microsoft Cortana, and he was chief scientist at Microsoft Health. Ryen has authored hundreds of publications in areas such as information retrieval, computational health, and human-computer interaction - including many that received awards. He was program chair for SIGIR 2017 and The Web Conference 2019. Ryen is editor-in-chief of ACM Transactions on the Web.

\noindent
\paragraph{Maarten de Rijke} is Distinguished University Professor of Artificial Intelligence and Information Retrieval at the University of Amsterdam. He is VP Personalization and Relevance and Senior Research Fellow at Ahold Delhaize. His research strives to build intelligent technology to connect people to information. His team pushes the frontiers of search engines, recommender systems and conversational assistants. They also investigate the influence of the technology they develop on society. De Rijke is the director of the Innovation Center for Artificial Intelligence.

\noindent
\paragraph{Sharada Mohanty} is the CEO and Co-founder of AIcrowd, a community of AI researchers built around a platform encouraging open and reproducible artificial intelligence research. He was the co-organizer of many large-scale machine learning competitions, such as NeurIPS 2017: Learning to Run Challenge, NeurIPS 2018: AI for Prosthetics Challenge, NeurIPS 2018: Adversarial Vision Challenge, NeurIPS 2019: MineRL Competition, NeurIPS 2019: Disentanglement Challenge, NeurIPS 2019: REAL Robots Challenge, NeurIPS 2020: Flatland Competition, NeurIPS 2020: Procgen Competition. He is extremely passionate about benchmarks and building communities. During his Ph.D. at EPFL, he worked on numerous problems at the intersection of AI and health, with a strong interest in reinforcement learning. In his previous roles, he has worked at the Theoretical Physics department at CERN on crowdsourcing compute for PYTHIA powered Monte-Carlo simulations; he has had a brief stint at UNOSAT building GeoTag-X---a platform for crowdsourcing analysis of media coming out of disasters to assist in disaster relief efforts. In his current role, he focuses on building better engineering tools for AI researchers and making research in AI accessible to a larger community of engineers.

\subsection{Partners}

\paragraph{Microsoft Research} has a mission to support research in computer science and software engineering. Microsoft Research in collaboration with Minecraft has developed Malmo, which is a platform for artificial intelligence experimentation~\citep{johnson2016malmo}, that has stimulated rapid research in the area~\citep{shu2017hierarchical}. Microsoft will support this competition by providing:
\begin{itemize}[nosep]
    \item cloud computing through Azure for participants to train their models;
    \item continued development support of the distributed data collection tool to reach out to Minecraft players (especially valuable for Task 2);
    \item a budget and proper IRB process for consent for the data collection with human-to-human players; and
    \item support for the final evaluation of the developed agents by adding the human-in-the-loop setup.
\end{itemize}

\paragraph{Facebook AI Research} has a long-standing goal to develop interactive assistants that can learn from human interactions. In this respect, FAIR invested into the development of the CraftAssist framework\footnote{\url{https://github.com/facebookresearch/droidlet/tree/main/craftassist}} which has matured into droidlet\footnote{\url{https://github.com/facebookresearch/droidlet}} - that makes it easier to utilize crowd-sourcing platforms to collect the human-to-human and human-assistant interactions at scale. FAIR will support us with 
\begin{itemize}[nosep]
    \item tooling for data collection in crowd-sourcing platforms such as MTurk\footnote{\url{https://www.mturk.com/}};
    \item extending data collection by enriching our crowd-sourcing infrastructure (especially valuable for Task 1 where we need to have rich language interactions).
\end{itemize}

\noindent 
In general, our partners are dedicated to sponsoring a new dataset collection which will potentially open up numerous research directions, and arranging a final human-in-the-loop evaluation.

\subsection{Resources provided by organizers}

\subsection{Computational Resources}
Microsoft has a mission to democratize access
to AI will provide computational resources to the participants. The concrete amount is the subject of the current discussion, which will clear upon acceptance and a further number of participants requiring resources.

\subsection{Providing mentorship for the teams} Our organization and advisory boards are really passionate about bringing NLU/NLG and RL communities. Therefore, we are dedicated to provide mentorship to a number of teams. We will give preference to underrepresented groups and young researchers.

\subsection{Prizes}
We are currently discussing this with our partners. Moreover, we reached out to potential sponsors, such as Amazon and Google, who are potentially interested to help us. We can get more details and traction upon acceptance. 

\subsection{Support requested}
It would be great if NeurIPS could host a workshop (or an event) where our winning participants can share their results. As far as we understood from the call, there is no session planned for a competition track yet. We would like to ask for a number of conference registrations reserved for our winning participants.

\bibliographystyle{plainnat}
\bibliography{ref}

\end{document}